\documentclass[letterpaper]{article}

\usepackage{natbib,alifeconf}

\usepackage[ruled,linesnumbered,noend]{algorithm2e}
\usepackage{caption}
\usepackage{subcaption}
\usepackage{multirow}
\usepackage{todonotes}
\usepackage{float}
\usepackage{multicol} 
\usepackage{graphicx}
\usepackage{lipsum}
\usepackage{amssymb}
\usepackage{hyperref}

\title{Towards Self-Assembling Artificial Neural Networks through\\Neural Developmental Programs}

\author{Elias Najarro,  Shyam Sudhakaran, \& Sebastian Risi\\
IT University of Copenhagen, Denmark\\
\{enaj, shsu, sebr\}@itu.dk
}
\begin{document}
\maketitle

\begin{figure*}[h]
\centering
  \includegraphics[width=1\textwidth]{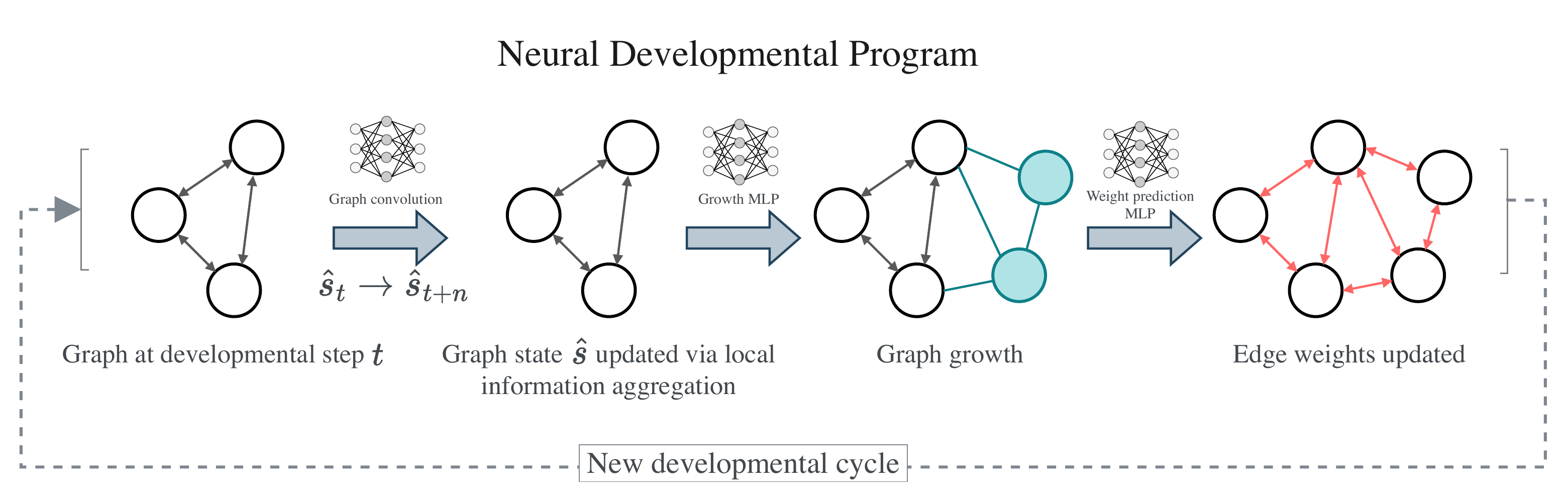}
  \caption{\textbf{Neural Developmental Program approach for growing neural network.} Each node state $s$ is represented as an embedding vector. During the information aggregation phase, the graph propagates each node state $s$ to its neighbors for $n$ steps. Based on the updated nodes embedding $\hat{s}_{t+n}$, the replication model —implemented as a MLP— determines which nodes will grow new nodes. Finally, if the target network is not unweighted, another MLP estimates the edge weights of each pair of nodes based on their concatenated embeddings; otherwise, edges are assigned a unitary weight. The resulting network is evaluated through an  objective function, i.e.\ solving a task or having certain topological properties. The $\mathcal{NDP}$ is a distributed model that operates purely on local information. }
  \label{fig:approach_overview}
\end{figure*}


\begin{abstract}
Biological nervous systems are created in a fundamentally different way than current artificial  neural networks. Despite its impressive results in a variety of different domains, deep learning often requires considerable engineering effort to design high-performing neural architectures. By contrast, biological nervous systems are grown through a dynamic self-organizing process. In this paper, we take initial steps toward neural networks that grow through a developmental process that mirrors key properties of embryonic development in biological organisms. The growth process is guided by another neural network, which we call a \emph{Neural Developmental Program} ($\mathcal{NDP}$) and which operates through local communication alone. We investigate the role of neural growth on different machine learning benchmarks and different optimization methods (evolutionary training, online RL, offline RL, and supervised learning). Additionally, we highlight future research directions and opportunities enabled by having self-organization driving the growth of neural networks.
\end{abstract}



%
\section{Introduction}

The study of neural networks has been a topic of great interest in the field of artificial intelligence due to their ability to perform complex computations with remarkable efficiency. However, despite significant advancements in the development of neural networks, the majority of them lack the ability to self-organize, grow, and adapt to new situations in the same way that biological neurons do. Instead, their structure is often hand-designed, and learning in these systems is restricted to the optimization of connection weights. 

Biological networks on the other hand, self-assemble and grow from an initial single cell. Additionally, the amount of information it takes to specify the wiring of a sophisticated biological brain directly is far greater than the information stored in the genome \citep{breedlove2013biological}. Instead of storing a specific configuration of synapses, the genome encodes a much smaller number of rules that 
govern how to grow a brain through a local and self-organizing process \citep{zador2019critique}. 
For example, the 100 trillion neural connections in the human brain are encoded by only around 30
thousand active genes. This outstanding compression has also been called the  ``genomic bottleneck'' \citep{zador2019critique}, and neuroscience suggests that this limited capacity has a regularizing effect that results in wiring and plasticity rules that generalize well.  

In this paper, we take first steps in investigating the role of developmental and self-organizing algorithms in growing neural networks instead of manually designing them, which is an underrepresented research area \citep{gruau1992genetic,nolfi1994phenotypic,kow2014growing,miller2014neuro}. Even simple models of development such as cellular automata demonstrate that  growth (i.e.\ unfolding of information over time) can be crucial to determining the final state of a system, which can not directly be calculated \citep{wolfram1984universality}. The grand vision is to create a  system in which neurons  self-assemble, grow, and adapt, based on the task at hand.

Towards this goal, we present a graph neural network type of encoding, in which the growth of a policy network (i.e.\ the neural network controlling the actions of an agent) is controlled by another network running in each neuron, which we call a \emph{Neural Developmental Program} ($\mathcal{NDP}$). The $\mathcal{NDP}$ takes as input information from the connected neurons in the policy network and decides if a neuron should replicate and how each connection in the network should set its weight. Starting from a single neuron, the approach grows a functional policy network, solely based on the local communication of neurons. Our approach is different from methods like NEAT \citep{stanley2002evolving} that grow neural networks during evolution, by growing networks during the lifetime of the agent. While not implemented in the current $\mathcal{NDP}$ version, this will ultimately allow the neural network of the agents to be shaped based on their experience and environment. 

While indirect genome-to-phenotype encodings such as CPPN-based approaches \citep{stanley2007compositional} or Hypernetworks \citep{ha2016hypernetworks} have had great success, they often purposely abstracted away development and the process of self-organizational growth. However, in nature, these abilities seem essential in enabling the remarkable robustness to perturbations and unexpected changes \citep{ha2021collective, risi2021selfassemblingAI}. Allowing each neuron in an artificial neural network to act as an autonomous agent in a decentralized way  similar to their biological counterpart \citep{Hiesinger2018Jul}, could enable our AI methods to overcome some of their current limitations in terms of robustness. 


Since the space of possible $\mathcal{NDP}$  representations is large, we explore two different representations and different training methods such as evolutionary and gradient-descent based. While lacking state-of-the-art performance, our method can learn to grow networks and policies that can perform competitively, opening up interesting future work in growing and developmental deep neural networks.  Our goal is to inspire researchers to explore the potential of $\mathcal{NDP}$-like methods as a new paradigm for self-assembling artificial neural networks. Overall, this work represents a step towards the development of more biologically inspired developmental encodings, which have the potential to overcome some of the limitations of current deep learning approaches.


\section{Background and Related Work}

\subsection{Indirect encodings} 

Indirect encodings are inspired by the biological  process of mapping a compact genotype to a larger phenotype and have been primarily studied in the context of neuroevolution \citep{floreanoDuerrMattiussi2008} (i.e.\ evolving  neural networks) but more recently were also optimized through gradient-descent based approaches \citep{ha2016hypernetworks}. In indirect encodings,  the description of the solution is compressed, allowing information to be reused and the final solution to contain more components than the description itself. Even before the success of deep RL, these methods enabled solving challenging car navigation tasks from pixels alone \citep{koutnik2013evolving}. 

A highly influential indirect encoding is  HyperNEAT~\citep{stanley2009hypercube}. HyperNEAT employs an indirect encoding called compositional pattern producing networks (CPPNs) that \emph{abstracts away the process of growth} and instead describes the connectivity of a neural network through a function of its geometry; an extension called Evolvable-substrate HyperNEAT \citep{risi2012enhanced} allowed neural architectures to be discovered automatically but did not involve any self-organizing process. More recently,  Hypernetworks \citep{ha2016hypernetworks} demonstrated that networks generating the weights of another network can also be trained end-to-end through gradient descent. Hypernetworks have been shown to be able to generate competitive convolutional neural networks (CNNs)  \citep{zhmoginov2022hypertransformer} and recurrent neural networks (RNNs) in a variety of tasks while using a smaller set of trainable parameters.  However, neither HyperNEAT nor Hypernetworks make use of the process of development over time, which can increase the evolvability of artificial agents \citep{kriegman2017minimal, bongard2011morphological} and is an important ingredient of biological systems \citep{hiesingerself}. 


\subsubsection{Developmental Encodings} Developmental encodings are a particular family of indirect encodings. They are abstractions of the developmental process that allowed nature to produce complex artifacts through a process of growth and local interactions between cells,  ranging  from  the  low  level  of  cell chemistry simulations to high-level grammatical rewrite systems \citep{stanley2003taxonomy}, and neurogenesis approaches \citep{Miller2022Feb, Maile2022Feb, Tran2022Nov, Huang2023Apr}. Approaches with neural networks that can grow  are a largely under-explored area \citep{gruau1992genetic,nolfi1994phenotypic,kow2014growing,miller2014neuro,richards2014} because these algorithms are either not expressive enough or not efficiently searchable.

Recently,  cellular automata (CA) had a resurgence of interest  as a model of biological development.  CA are a class of computational models whose outcomes emerge from the local interactions of simple rules. Introduced by  \citet{neumann1966theory} as a part of his quest to build a self-replicating machine or universal constructor, a CA consist of a lattice of computing cells that iteratively update their states based exclusively on their own state and the states of their local neighbors. On a classical CA, each cell's state is represented by an integer and adjacent cells are considered neighbors. Critically, the update rule of a cellular automaton is identical for all the cells.

 Neural cellular automata (NCA) differ from classical cellular automata (CA) models by replacing the CA update function with an optimized neural network \citep{mordvintsev2020growing,nichele2017neat}. Recently, this approach has been extended to grow complex 3D entities such as castles, apartment blocks, and trees in a video game environment \citep{sudhakaran2021growing}. 

A recent method called HyperNCA, extends NCA to grow a 3D pattern, which is then mapped to the weight matrix of a policy network \citep{Najarro2022Apr}. While working well in different reinforcement learning tasks, the mapping from the grown 3D pattern to policy weight matrix did not take the topology of the network into account. 
Instead of a  grid-like HyperNCA approach, the method presented in this paper extends NCA to directly operate on the policy graph itself and should thus also allow more flexibility in the types of architectures that can be grown.  

\subsection{Distribution-fitting approaches to evolving graphs} Previous work has explored the emerging topologies of the different growth processes \citep{Albert2000Dec} and shown that they can reproduce real networks by fitting the parameters of the distribution from which new nodes are sampled. In contrast to our method, the growth processes in previous network-theory approaches
do not depend on the internal state of the graph, and therefore do not make use of the developmental aspect of the network to achieve the target topological properties.

\section{Approach: Growing Neural Networks through Neural Developmental Programs}
This section presents the two different Neural Developmental Program instantiations we are exploring in this paper: (1) an evolution-based $\mathcal{NDP}$ and (2) a differentiable version trained with gradient descent-based. While an evolutionary version allows us to more easily explore different architectures without having to worry about their differentiability,  gradient descent-based architectures can often be more sample efficient, allow scaling to higher dimensions, and enable approaches such as offline reinforcement learning. Code implementations will be available soon at:  \url{https://github.com/enajx/NDP}.


\subsection{Evolutionary-based $\mathcal{NDP}$}

The  $\mathcal{NDP}$ consists of a series of developmental cycles applied to an initial seeding graph; our experiments always use a seeding graph consisting of a single node or a minimal network connecting the neural network's inputs directly to its outputs. Each of the nodes of the graph has an internal state represented as  $n-$dimensional latent vector whose values are updated during the developmental process through local communication. The node state-vectors —or embeddings— encode the cells' states and are used by the $\mathcal{NDP}$ to determine which nodes will duplicate to make new nodes.

Similarly to how most cells in biological organisms contain the same program in the form of DNA, each node's growth and the synaptic weights are controlled by a copy of the same $\mathcal{NDP}$, resulting in a distributed self-organized process that incentives the reuse of information. An overview of the approach is shown in Fig.~\ref{fig:approach_overview}.

The $\mathcal{NDP}$ architecture consists of a Multilayer Perceptron (MLP) —acting as a Graph Cellular Automata (GNCA) \citep{Grattarola2021Oct}— which updates the node embeddings after each message-passing step during the developmental phase. Subsequently, a replication model in the form of a second MLP queries each node state and predicts whether a new node should be added; if so, a new node is connected to the parent node and its immediate neighbors. Finally, if the target network is weighted, a third MLP determines the edge weights based on the concatenation of each pair of node embeddings. The grown network can now be evaluated on the task at hand by assigning a subset of nodes as the input nodes and another subset as the output nodes. 
In our case, we select the first —and last— rows of the adjacency matrix representing the network to act as input — and output— nodes, respectively. During evaluation the activations of the nodes are scalars (that is, $\mathbb{R}^1$ instead of the $\mathbb{R}^n$ vectors used during development), and all node activations are initialized to zero.

We refer to the $\mathcal{NDP}$ as the set of these MLPs which are identical for each cell in the policy network; in order to keep the number of parameters of the $\mathcal{NDP}$ low, the reported experiments make use of small MLPs with a single hidden layer. However, it's worth noticing that because the $\mathcal{NDP}$  is a distributed model (i.e.\  the same models are being applied to every node), the number of parameters is constant with respect to the size of the graph in which it operates. Therefore, any neural network of arbitrary size or architecture  could be used, provided that it was deemed necessary to grow a more complex graph. The $\mathcal{NDP}$'s neural networks can be trained with any black-box optimization algorithm to satisfy any objective function. In this paper, we demonstrate how the approach allows to grow neural networks capable of solving reinforcement learning and classification tasks, or exhibiting some topological properties such as small-worldness. The pseudocode of the approach is detailed in Algorithm~\ref{algo1}.

\begin{algorithm}[t]
\SetAlgoLined
\vspace{1mm}
\KwIn{Replication model $\mathcal{R}$, Graph Cellular Automata $\mathcal{GNCA}$, Weight update model $\mathcal{W}$, number of developmental cycles $\mathcal{C}$, pruning threshold $\mathcal{P}$, number of training generations $\mathcal{T}$, training hyper-parameters $\Omega$}
\KwOut{Developmental program producing graph $\mathcal{G}$ that minimise/maximise $\mathcal{F}$} 
\vspace{1mm}
Co-evolve or sample random embedding $E_{N=0}$ for the initial node $N_0$ of $\mathcal{G}$; \\
\For(){generation \textbf{in} $\mathcal{T}$}
    {
    \For(){developmental cycle \textbf{in} $\mathcal{C}$}
        {
    
        Compute network diameter $\mathtt{D}$; \\
    
        Propagate nodes states $E_{N}$ via graph convolution  $\mathtt{D}$ steps; \\
    
        Replication model $\mathcal{R}$ determines nodes in growing state; \\
    
        New nodes are added to each of the growing nodes and their immediate neighbors; \\
    
        New nodes' embeddings are defined as the mean embedding of their parent nodes.
    
        \If{weighted network}{
        Weight update model $\mathcal{W}$ updates connectivity for each pair of nodes based on their concatenated embeddings\; 
        }
        \If{pruning}{
        Edges with weights below pruning threshold $\mathcal{P}$ are removed;\\ 
        }
    
        }
        
        Evaluate objective $\mathcal{F}$ of grown graph $\mathcal{G}$; \\
    
        Use $\mathcal{F}(G)$ to guide optimisation; \\
    }

 \caption{Neural Developmental Program $\mathcal{NDP}$: non-differentiable version }
 \label{algo1}
\end{algorithm}

\subsection{Gradient-based $\mathcal{NDP}$}

\begin{algorithm}[ht!]
\SetAlgoLined
\vspace{1mm}
\KwIn{Developmental model $\mathcal{D}$, number of developmental cycles $\mathcal{N_{D}}$, number of training iterations $\mathcal{T}$, replication model $\mathcal{R}$, edge prediction model $\mathcal{E}$, reinforcement learning algorithm $\mathcal{RL}$}
\KwOut{Developmental program producing graph $\mathcal{G}$ that maximises reward $\mathcal{F}$} 
\vspace{1mm}
Initialize trainable node embeddings $N_{init}$ of $\mathcal{G}$; \\
\For(){training iteration \textbf{in} $\mathcal{T}$}
    {
    \For(){i \textbf{in} $\mathcal{N_{D}}$}
        {

        1. Get new node embeddings $N_{i}$ via Neural Message Passing $\mathcal{D}(N_{initial})$ \\
    
        2. Sample parent node $\mathcal{P}_{i}$ using replication channel probabilities in node embeddings $N_{i}$

        3. Create replicated child node with replication model $\mathcal{C}_{i} = \mathcal{R}(\mathcal{P}_{i})$ and connect an incoming edge from $\mathcal{P}_{i}$ to $\mathcal{C}_{i}$. Connect an outgoing edge from $\mathcal{C}_{i}$ to a random node at the further on in the network (layers deeper in the network).
    
        4. Add child node to the network
    
        5. Predict edge weights using Edge prediction model $\mathtt{E}$ for each edge in the network, using pairs of source and destination node embeddings
    
        }
    Collect trajectories with a rollout using grown graph $\mathcal{G}$; \\

    Update parameters via RL algorithm $\mathcal{RL}$
    }

 \caption{Neural Developmental Program $\mathcal{NDP}$: differentiable version}
 \label{algo:diff_ndp}
\end{algorithm}

The gradient-based growth process is similar to the evolutionary approach, albeit with some additional constraints due to its requirement of complete differentiability in the process.
In contrast to the evolutionary approach, the grown networks are exclusively feedforward networks, where information is iteratively transmitted through message passing via the topologically sorted nodes. Each node has a bias value, and an activation that is applied to when the incoming nodes' information is aggregated, similar to the node behavior in e.g.\ NEAT \citep{stanley2002evolving}.

Like in the evolutionary approach, cells' states are represented as vectors. However, instead of all values of each cell's state vector being treated as intractable black-boxes for the network to store information, here each first and second elements of the vectors have pre-defined roles: the first element represents the bias value and the second  encodes the activation. The remaining encode hidden states for each cell, capturing temporal information used to guide the developmental process. These cells are passed into a message-passing network, implemented as a graph Convolution \citep{graphconv}, where a neighborhood of cell embeddings, characterized as being 1 edge away are linearly projected and added to create new embeddings. In addition to using a message passing network to update nodes, we use a trainable small 2-layer MLP, with tanh activation, to predict edge weights using pairs of (source, destination) nodes. The message-passing network is composed of a GraphConv layer that outputs vectors of size 32, followed by a Tanh activation, followed by a linear layer that maps the size 
 vectors to the original cell embeddings. 

In order to add more nodes, we treat a channel in each cell as a \emph{replication probability}, which is used to sample cells and connect to a random node further into the network. This process happens every other growth step.


Detailed replication process:
(1) A replication network, implemented as a separate graph convolution, is applied on cells to output replication probabilities. (2) A cell is sampled from these probabilities and is passed into a perturbation network to get a new cell, and an incoming edge is connected from the parent to the new cell. An outgoing edge is connected to a random child (found further in the network). After the new node is added to the network, we update the edges using an MLP that takes in incoming node and outgoing node pairs and outputs a new edge weight. 

We initialize the starting network by fully connecting trainable input cell embeddings and output cell embeddings, which we found to work better for our gradient-based $\mathcal{NDP}$ than starting with a single node.  For example, if input = 4 and output = 2, then each input node will be connected to an output node, resulting in 4$\times$2 initial edges. The pseudocode of this approach is detailed in Algorithm~\ref{algo:diff_ndp}.

 \begin{figure*}
      \centering
      \begin{subfigure} {0.20\textwidth}
      \includegraphics[width=1.2\textwidth]{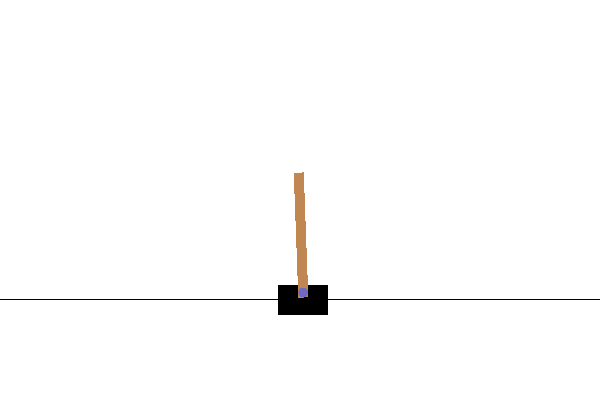}
      \centering
           \caption{CartPole}
          \label{fig:y equals x}
      \end{subfigure}
                  \begin{subfigure}{0.25\textwidth}
                        \centering
      \includegraphics[width=0.7\textwidth]{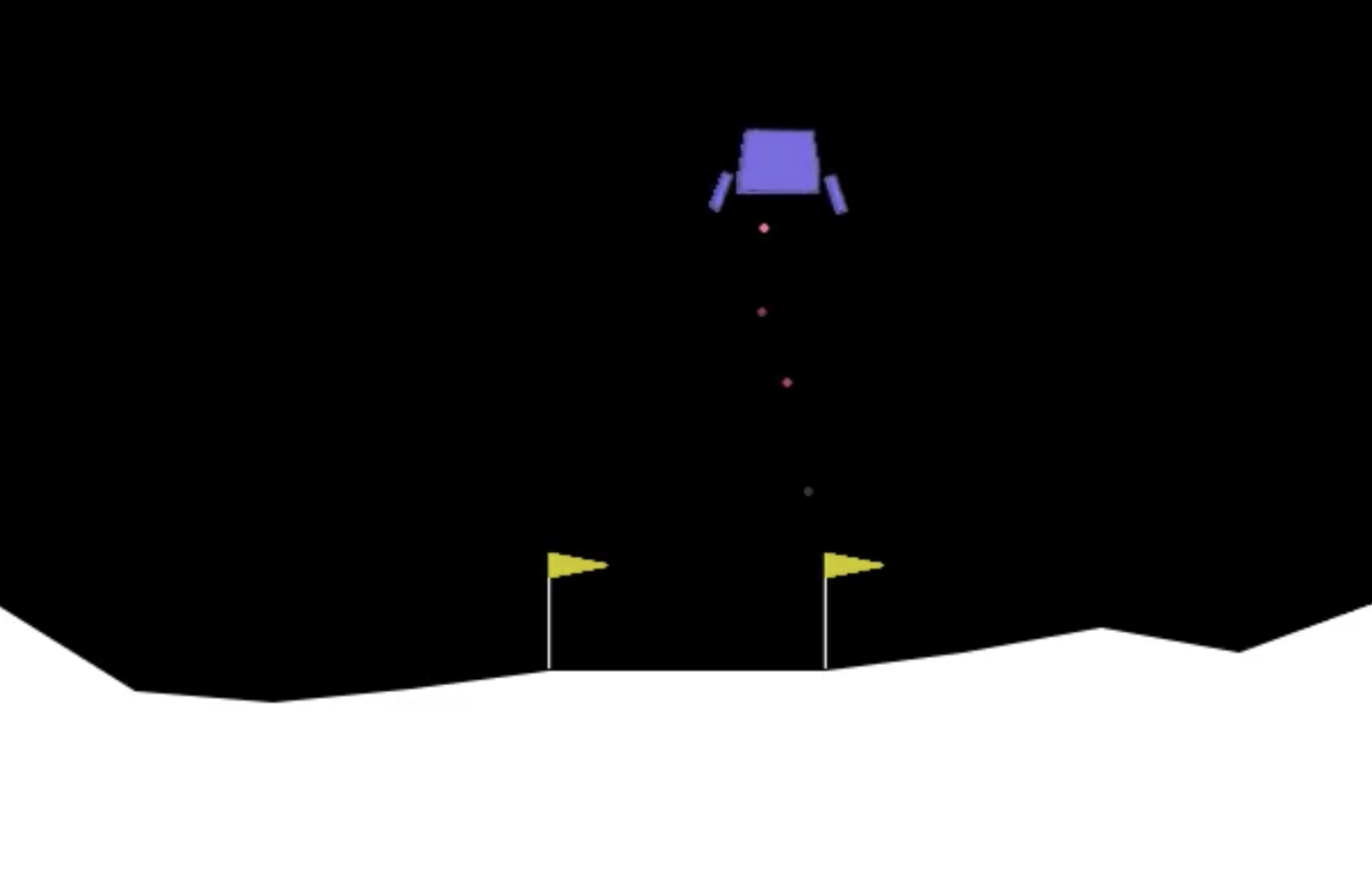}
           \caption{LunarLander}
          \label{fig:y equals x}
      \end{subfigure}
                        \begin{subfigure}{0.25\textwidth}
                              \centering
      \includegraphics[width=0.5\textwidth]{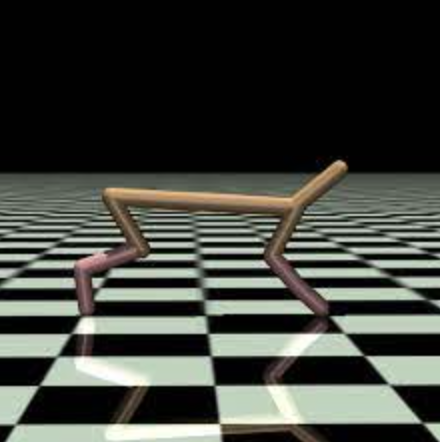}
      \caption{HalfCheetah}
          \label{fig:y equals x}
      \end{subfigure}
                  \begin{subfigure}{0.20\textwidth}
                  \centering
      \includegraphics[width=0.6\textwidth]{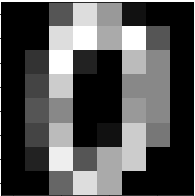}
           \caption{8x8 MNIST Example}
          \label{fig:y equals x}
      \end{subfigure}
      \caption{Test domains in this paper include both reinforcement and supervised tasks.}
      \label{fig:testdomains}
\end{figure*}

\section{Experiments}

We test the $\mathcal{NDP}$ approach on generating networks for classification tasks such as \emph{MNIST}, boolean gates such as \emph{XOR} gate and reinforcement learning tasks with both continuous and discrete action spaces. The RL tasks include \emph{CartPole}, a classical control task with non-linear dynamics, and \emph{LunarLander}, a discrete task where the goal is to smoothly land on a procedurally generated terrain. We also tackle an offline RL task using Behavioral Cloning (BC) \citep{bc}, which is  \emph{HalfCheetah}, a canine-like robot task.

The different environments are shown in Fig.~\ref{fig:testdomains}. The \emph{CartPole} environment provides observations with 4 dimensions, \emph{LunarLander} has 8, MNIST has 64, and HalfCheetah has 17. \emph{CartPole} has 2 actions, \emph{LunarLander} has 4 actions, \emph{MNIST} has 10 classes to predict, and \emph{HalfCheetah} has a continuous action of dimension 7. 




\subsection{Evolutionary Optimization Details}

We use CMA-ES — Covariance Matrix Adaptation Evolution Strategy — \citep{Hansen1996May}, a black-box population-based optimisation algorithm to train the $\mathcal{NDP}$. Evolutionary strategies (ES) algorithms have been shown to be capable of finding high-performing solutions for reinforcement learning tasks, both directly optimising the policy weights \citep{Salimans2017Mar}, and with developmental encodings  \citep{Najarro2022Apr}. Black-box methods such as CMA-ES have the advantage of not requiring to compute gradients and being easily parallelizable. 

Experiments have been run on a single machine with a \textit{AMD Ryzen Threadripper 3990X} CPU with $64$ cores. We choose a population size of $512$, and an initial variance of $0.1$. Finally, we employ an early stopping method which stops and resets training runs that show poor performance after a few hundred generations.

\subsection{Differentiable Optimization Details}

We use the Pytorch Geometric library to implement our $\mathcal{NDP}$ as well as our grown networks, enabling us to backpropagate a loss using predictions from the grown networks all the way back to the parameters of the $\mathcal{NDP}$. We are also able to leverage GPUs for the forward and backward passes.

Experiments have been run on a single machine with a \textit{NVIDIA 2080ti} GPU. We use the Adam Optimizer \citep{adam} to optimize the $\mathcal{NDP}$'s trainable parameters.

\subsubsection{Online RL with PPO}
Because we can take advantage of backpropagation with the differentiable $\mathcal{NDP}$ approach, we can utilize reinforcement learning algorithms, specifically Proximal Policy Optimization (PPO) \citep{ppo}, to grow optimal policies. Implementations of PPO typically use separate networks / shared networks for critic and actor heads. In our implementations, we simply treat the critic output as a separate node that is initially fully connected to all the input nodes. In our implementation, we use a learning rate of 0.0005, an entropy coefficient of 0.001, and optimize for 30 steps after each rollout is collected. We train for 10,000 rollouts and record the average reward from 10 episodes in Tables~\ref{tab:cartpole_grad} and \ref{tab:lunar_grad}.

\subsubsection{Offline RL with Behavioral Cloning}
In Offline RL, instead of gathering data from an environment, we only have access to a dataset of trajectories. This setting is challenging but also easier because we can avoid the noisy training process of Online RL. In our approach, we utilize Behavioral Cloning (BC) to optimize our $\mathcal{NDP}$ to grow high-performing policies. We use a learning rate of 0.0001 and a batch size of 32 observations. We train for 10000 rollouts and record the average reward from 10 episodes in Table~\ref{tab:haflcheetah_grad}.

\subsubsection{Supervised Learning}
We evaluate the supervised learning capabilities of our differentiable $\mathcal{NDP}$ with the classic \emph{MNIST} task, where a small (8$\times$8) image is classified as a digit between 0-9. We use a learning rate of 0.001 and a batch size of 32 observations. We train for 10000 iterations and record the test accuracy in Table~\ref{tab:mnist_grad}.

\section{Evolutionary-based training results}

\begin{figure}
  \includegraphics[width=0.47\textwidth]{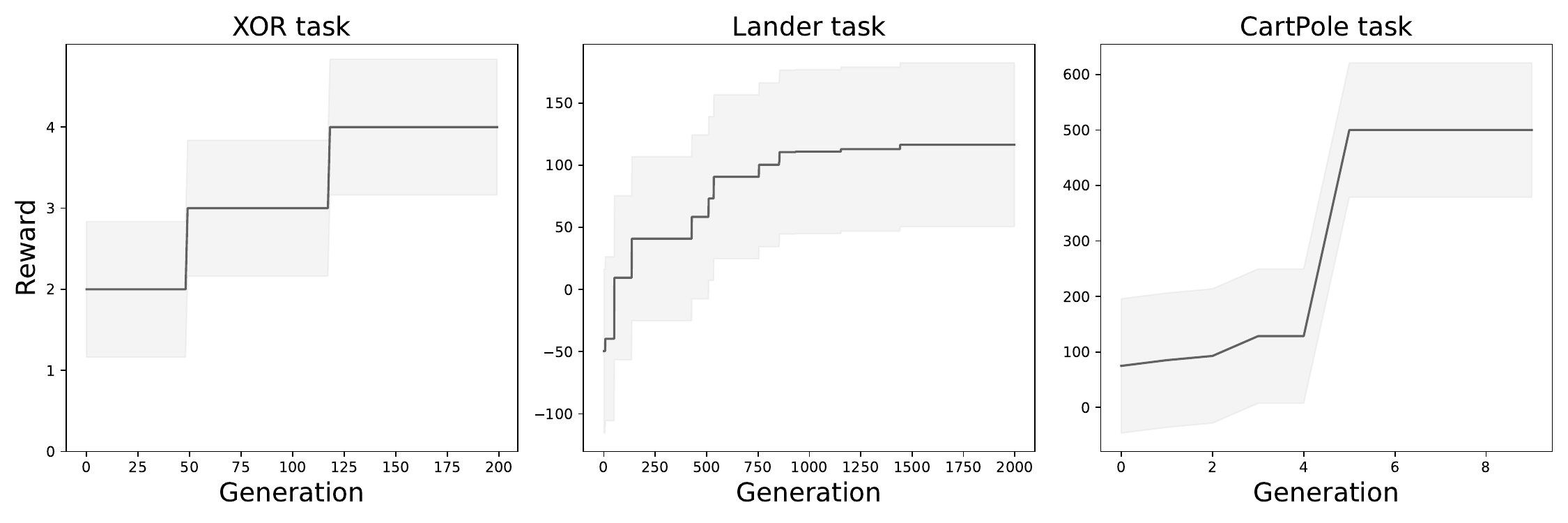}
  \caption{Fitness during evolution on the XOR, LunarLander and CartPole tasks. Gray areas  show standard deviation at each generation of the CMA-ES sampled solutions. }
  \label{fig:fitnesses}
\end{figure}

\begin{figure}[t]
  \centering
  \includegraphics[width=0.43\textwidth]{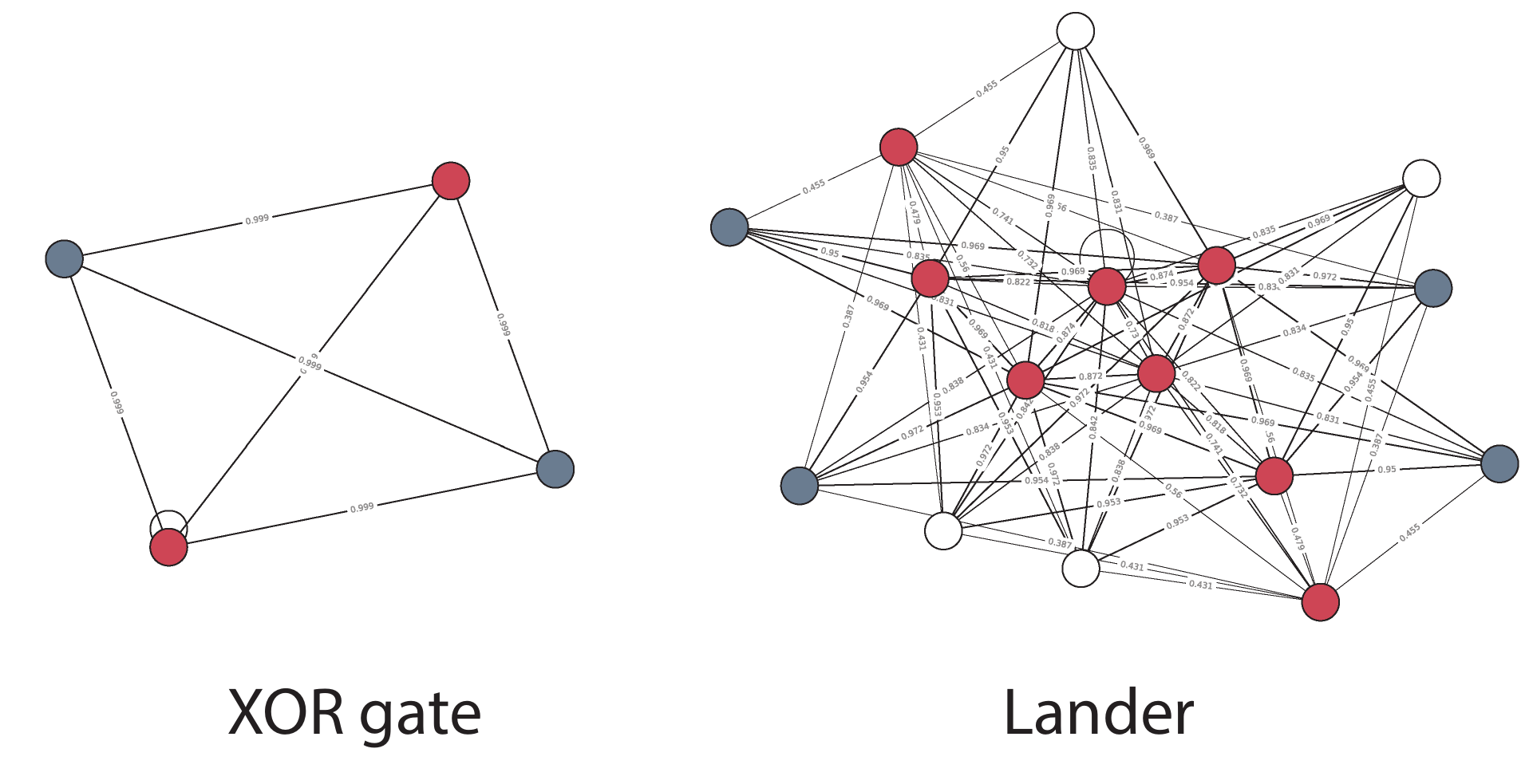}
  \caption{Grown networks solving XOR gate task (left), and Lunar Lander task (right). Red nodes behave as sensory neurons, white nodes are interneurons, and blue ones are the \textit{output} neurons that determine the actions of the cartpole. }
  \label{fig:XORandLander}
\end{figure}

\textbf{Growing networks for XOR gate.} 
The XOR gate is a classic problem, which interests reside in its non-linearly separable nature. While the $\mathcal{NDP}$ has in fact more parameters than necessary to solve the simple XOR task (which can be solved with a neural network with one hidden  node), it serves as the initial test that networks with multiple layers can be grown with the evolutionary $\mathcal{NDP}$. Indeed, we find that a simple $\mathcal{NDP}$ with 14 trainable parameters can grow an undirected graph capable of solving the XOR gate  (Fig.~\ref{fig:XORandLander}, left). The resulting graph has 4 nodes and 7 weighted edges. Graph layouts are  generated using the Fruchterman-Reingold force-directed algorithm.  Training curves are shown in Fig.~\ref{fig:fitnesses}. 


\begin{figure*}
  \includegraphics[width=1\textwidth]{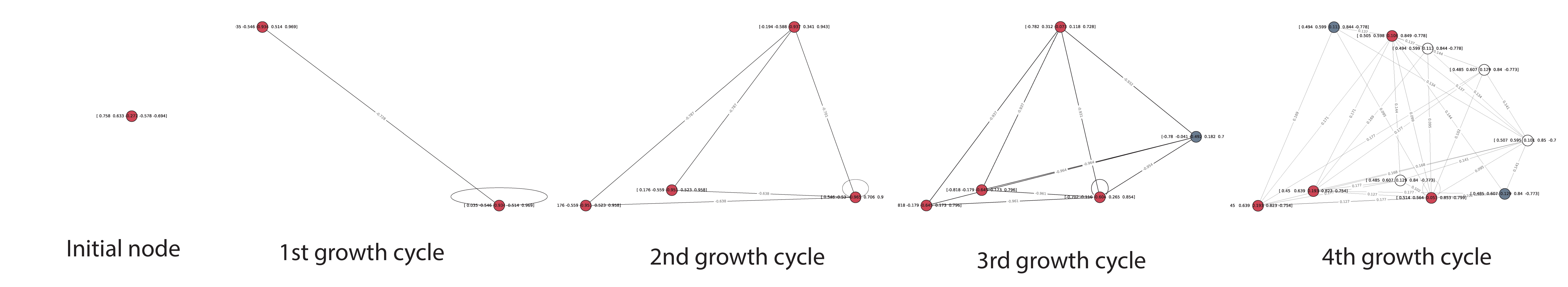}
  \caption{Developmental growth of the network capable of solving the Cart Pole balancing task. The network starts as a single node and grows to a network of size 2, 4, 5, and finally 10 neurons and 33 weighted edges after 4 growth cycles. Red nodes behave as sensory neurons, white nodes as interneurons, and blue ones act as the \textit{output} neurons that determine the actions of the cartpole. The vector displayed above the neurons are the node embeddings which represent the state of each node during the growth process.}
  \label{fig:cartpolegrow}
\end{figure*}

\textbf{Growing policy network for RL tasks.} We trained a $\mathcal{NDP}$ with 162 parameters  for the CartPole tasks, in which it has to grow a policy network controlling the force applied to a pole to keep it balanced. It grew an undirected network with 10 nodes and 33 weighted edges that reached a reward of $500 \pm 0$  over 100 rollouts (Fig.~\ref{fig:fitnesses}). This score is considered as solving the task. The growth process of the network from a single node to the final policy can be seen in Fig.~\ref{fig:cartpolegrow}.

Similarly, we trained a $\mathcal{NDP}$ to grow a network policy to solve the Lunar Lander control task (Fig.~\ref{fig:XORandLander}, right). In this case, a $\mathcal{NDP}$ with 868 trainable parameters grew an undirected network policy with 16 nodes and 78 weighted edges. Over 100 rollouts, the mean reward obtained is $116 \pm 124$. Although the resulting policy controller could solve the task in many of the rollouts, the stochasticity of the environment (e.g.\ changing the landing surface at each instantiation) resulted in a high reward variance. This means that the grown network did not quite reach the 200 reward over 100 rollouts score that is considered as the task's solving criterion.

\textbf{Growing small-world topologies.} 
Real networks found in nature such as biological neural networks, ecological networks or social networks are not simple random networks but rather their graphs have very specific topological properties.  In order to analyze the ability of the neural developmental program to grow complex graph motifs, we train the $\mathcal{NDP}$ to grow a small-world network \citep{Watts1998Jun}. A small-world network is characterised by a small average shortest path length but a relatively large clustering coefficient. We can quantify this with two metrics $\sigma = \frac{C/C_{r}}{L/L_{r}}$ and $\omega = \frac{L_{r}}{L} - \frac{C}{C_{l}}$, where $C$ and $L$ are respectively the average clustering coefficient and average shortest path length of the network. $C_{r}$ and $L_{r}$ are respectively the average clustering coefficient and average shortest path length of an equivalent random graph. A graph is commonly classified as small-world if $\sigma>1$ or, equivalently, $\omega \approx 0$.

\begin{figure}[h]
\centering
  \includegraphics[width=0.30\textwidth]{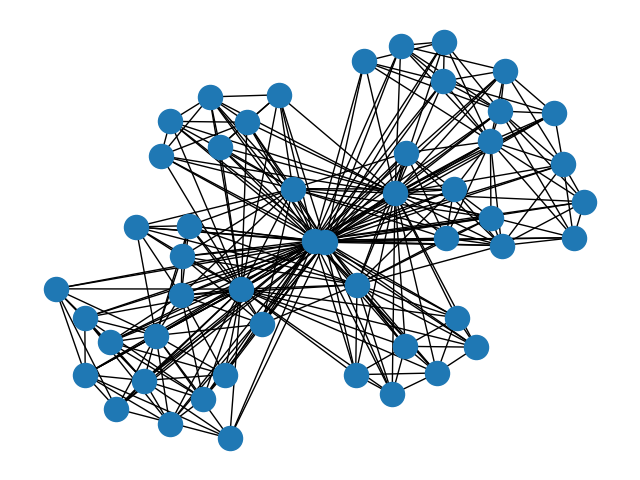}
  \caption{Grown Small-World network. In this experiment, the $\mathcal{NDP}$ seeks to maximise the coefficients used to capture the small-worldness criteria of a graph. Unlike the networks grown to act as policies, this network is unweighted. }
  \label{fig:SmallWorld}
\end{figure}

We show that $\mathcal{NDP}$ technique can grow a graph with small-world coefficients are $\sigma\approx1.27$ and $\omega\approx-1.11^{-16}$, hence satisfying the small-worldness criteria. An example network is shown in Fig.~\ref{fig:SmallWorld}. While these results are promising, further experiments are required —notably on bigger graphs— to investigate with solid statistical significance the potential of the method to grow graphs with arbitrary topological properties.

\section{Gradient-Based Results}


\begin{table*}[h]
    \begin{subtable}[h]{0.45\linewidth}
        \centering
        \begin{tabular}{|l | l | l|}
        \hline
        Growth Steps & Rew. & Network Size \\
        \hline \hline
         1  & $123 \pm 11.3$ & 7  \\
         12 & $257 \pm 6.9$ & 13  \\
         24 & $334 \pm 4.4$ & 19 \\
         48 & $500 \pm 3.1$ & 31 \\
         64 & $500 \pm 1.7$ & 39  \\
         128 & $500 \pm 2.8$ & 71 \\
        \hline
       \end{tabular}
       \caption{CartPole Results. Number of $\mathcal{NDP}$ parameters: 11223.}
       \label{tab:cartpole_grad}
    \end{subtable}
    \hfill
    \begin{subtable}[h]{0.45\linewidth}
        \centering
        \begin{tabular}{|l | l | l|}
        \hline
        Growth Steps & Rew. & Network Size \\
        \hline \hline
         1  & $24\pm 7.3$ & 13\\
         12 & $68\pm 8.1$ & 19 \\
         24 & $100\pm 8.7$ & 25 \\
         48 & $130\pm 3.4$ & 37 \\
         64 & $112\pm 5.8$ & 45 \\
         128 & $110\pm 6.4$ & 77 \\
        \hline
       \end{tabular}
       \caption{LunarLander Results. Number of $\mathcal{NDP}$ parameters: 11445.}
       \label{tab:lunar_grad}
    \end{subtable}
     \caption{\textbf{Differentiable NDP}: Online RL Results. Mean reward is calculated over 10 test episodes after 10,000 rollouts were collected. Networks are trained with a specific number of growth steps, as shown in the ``growth steps'' column.}
     \label{tab:onlineRL}
\end{table*}

\begin{table*}[h]
    \begin{subtable}[h]{0.45\linewidth}
        \centering
        \begin{tabular}{|l | l | l|}
        \hline
        Growth Steps & Test Acc. & Network Size \\
        \hline \hline
         1  & $13 \pm 8.1$ & 74 \\
         12 & $33 \pm 6.3$ & 80 \\
         24 & $78 \pm 4.7$ & 86 \\
         48 & $93 \pm 2.9$ & 98 \\
         64 & $91 \pm 1.4$ & 106 \\
        \hline
       \end{tabular}
       \caption{MNIST Results. Number of $\mathcal{NDP}$ parameters: 13739.}
       \label{tab:mnist_grad}
    \end{subtable}
    \hfill
    \begin{subtable}[h]{0.45\linewidth}
        \centering
        \begin{tabular}{|l | l | l|}
        \hline
        Growth Steps & Rew. & Network Size \\
        \hline \hline
         1  & $13 \pm 2.2$ & 24 \\
         12 & $17 \pm 2.6$ & 30 \\
         24 & $23 \pm 2.3$ & 36 \\
         48 & $25 \pm 1.7$ & 48 \\
         64 & $29 \pm 1.1$ & 56 \\
        \hline
       \end{tabular}
       \caption{HalfCheetah Results. Number of $\mathcal{NDP}$ parameters: 11889.}
       \label{tab:haflcheetah_grad}
    \end{subtable}
     \caption{\textbf{Differentiable $\mathcal{NDP}$}: Supervised Learning and Offline RL tasks. Test accuracy for MNIST calculated after 10,000 epochs. Mean reward for HalfCheetah calculated over 10 test episodes after 10000 epochs.}
     \label{tab:offlineRL}
\end{table*}





\begin{figure*}[h]
     \centering
     \begin{subfigure}[b]{0.19\textwidth}
         \centering
         \captionsetup{justification=centering}
         \includegraphics[width=\textwidth]{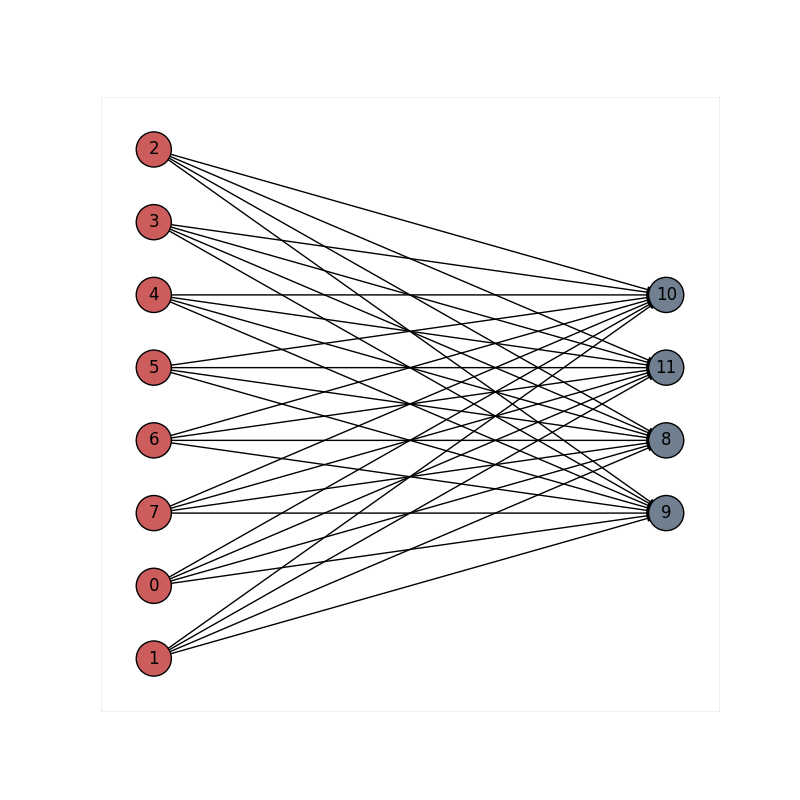}
         \caption*{Initial seeding graph. \\13 Nodes.}
         \label{fig:y equals x}
     \end{subfigure}
     \hfill
     \begin{subfigure}[b]{0.19\textwidth}
         \centering
         \captionsetup{justification=centering}
         \includegraphics[width=\textwidth]{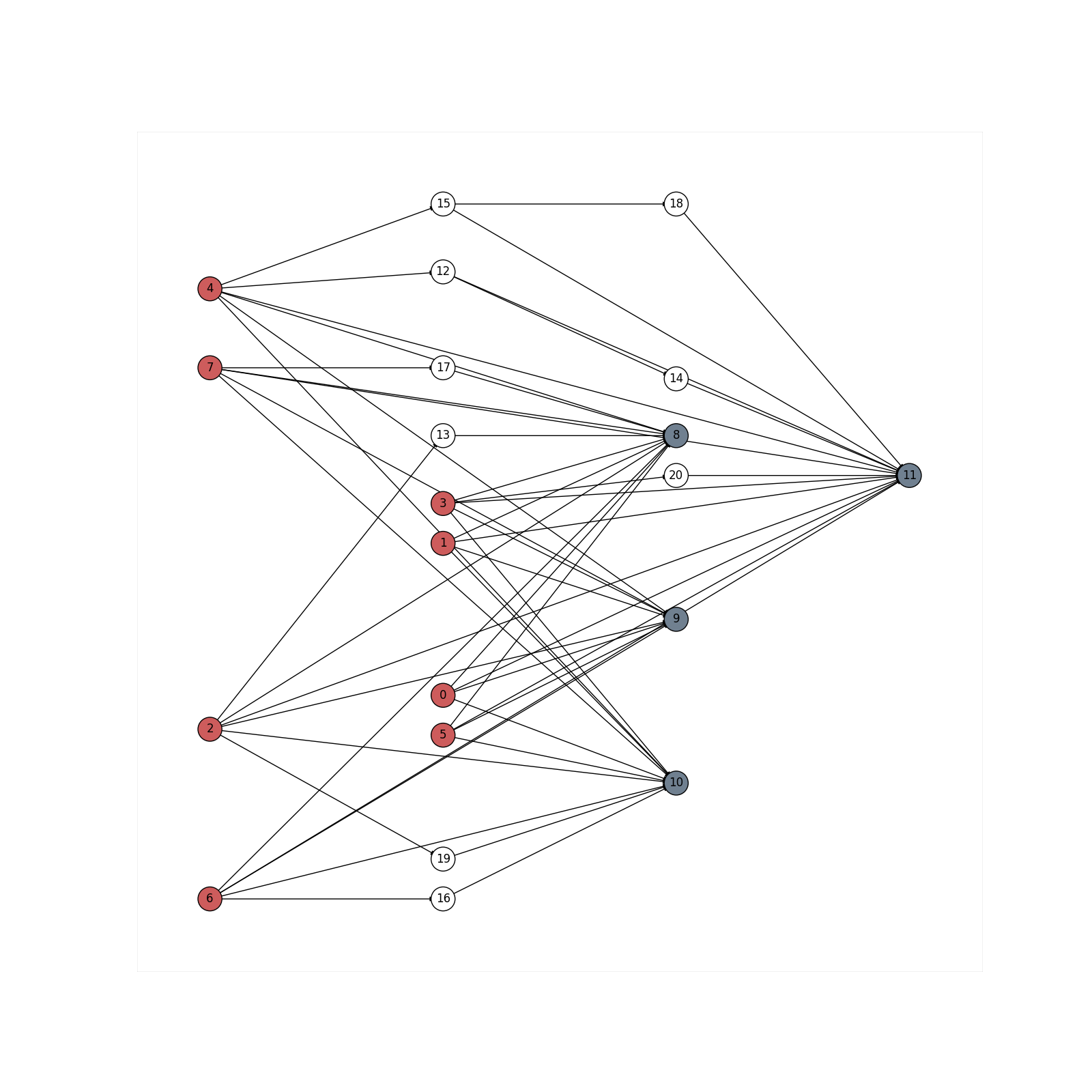}
         \caption*{16 growth steps.\\ 21 Nodes.}
         \label{fig:three sin x}
     \end{subfigure}
     \hfill
     \begin{subfigure}[b]{0.19\textwidth}
         \centering
         \captionsetup{justification=centering}
         \includegraphics[width=\textwidth]{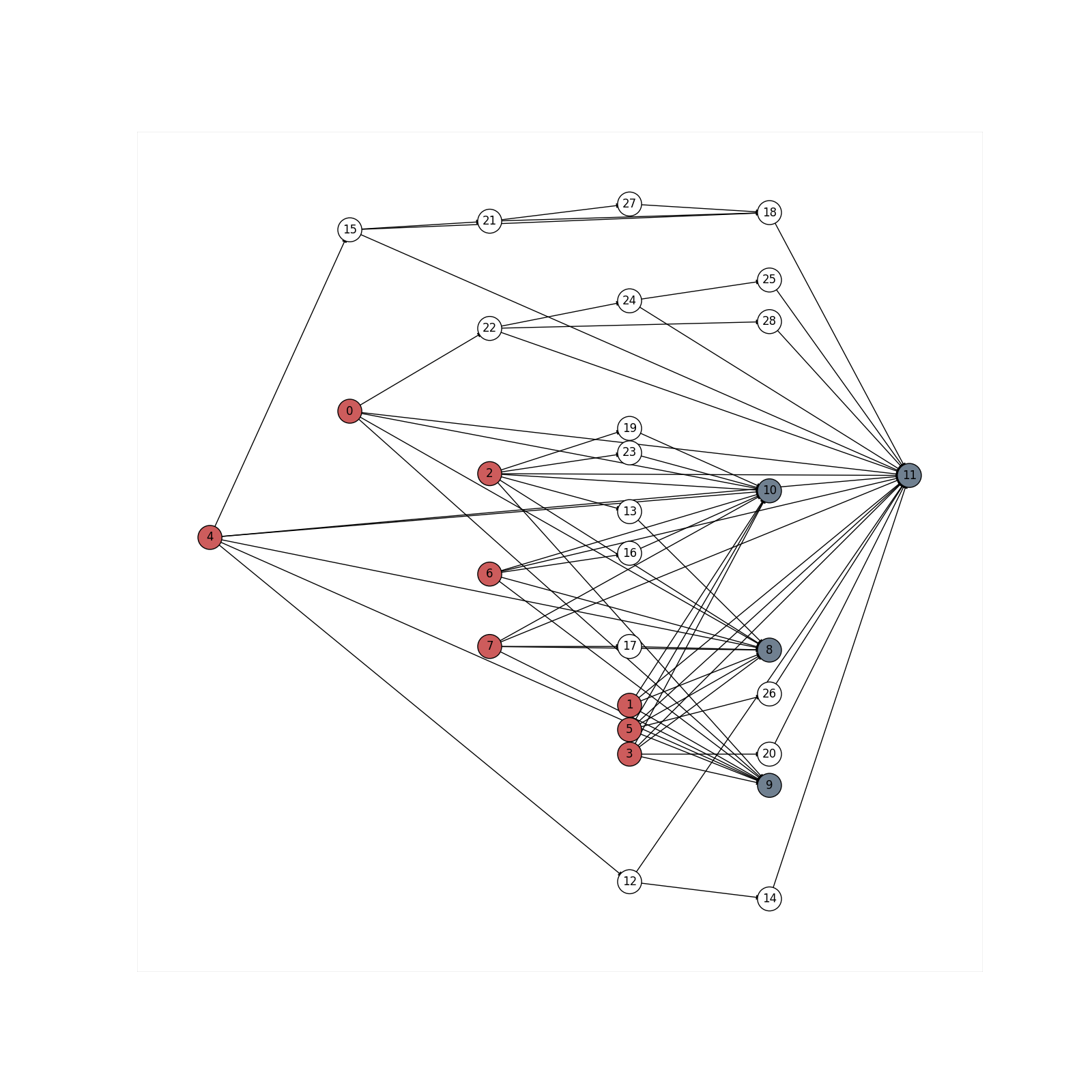}
         \caption*{32 growth steps. \\29 Nodes.}
         \label{fig:five over x}
     \end{subfigure}
     \hfill
     \begin{subfigure}[b]{0.19\textwidth}
         \centering
         \captionsetup{justification=centering}
         \includegraphics[width=\textwidth]{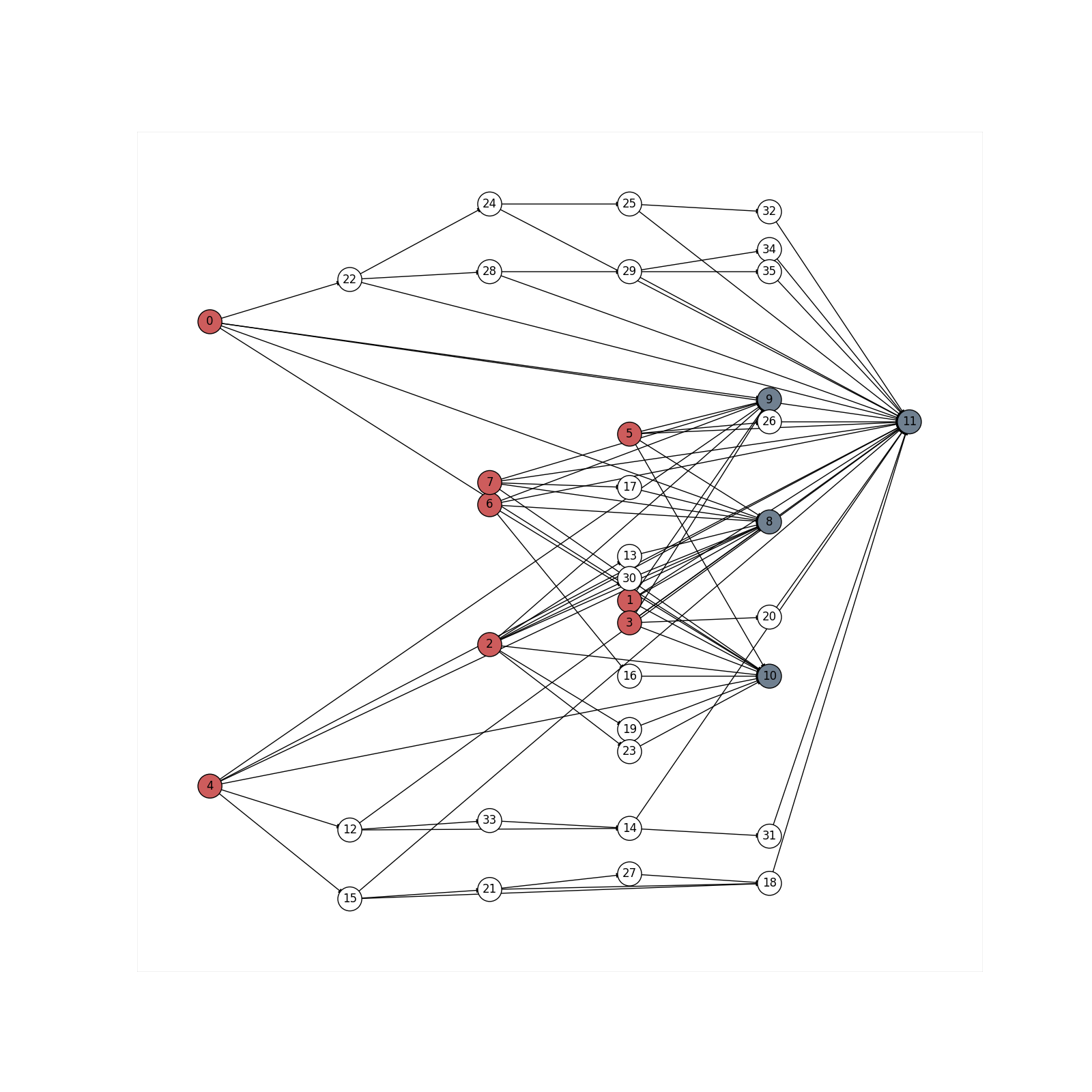}
         \caption*{48 growth steps.\\ 37 Nodes.}
         \label{fig:five over x}
     \end{subfigure}
     \hfill
     \begin{subfigure}[b]{0.19\textwidth}
         \centering
         \captionsetup{justification=centering}
         \includegraphics[width=\textwidth]{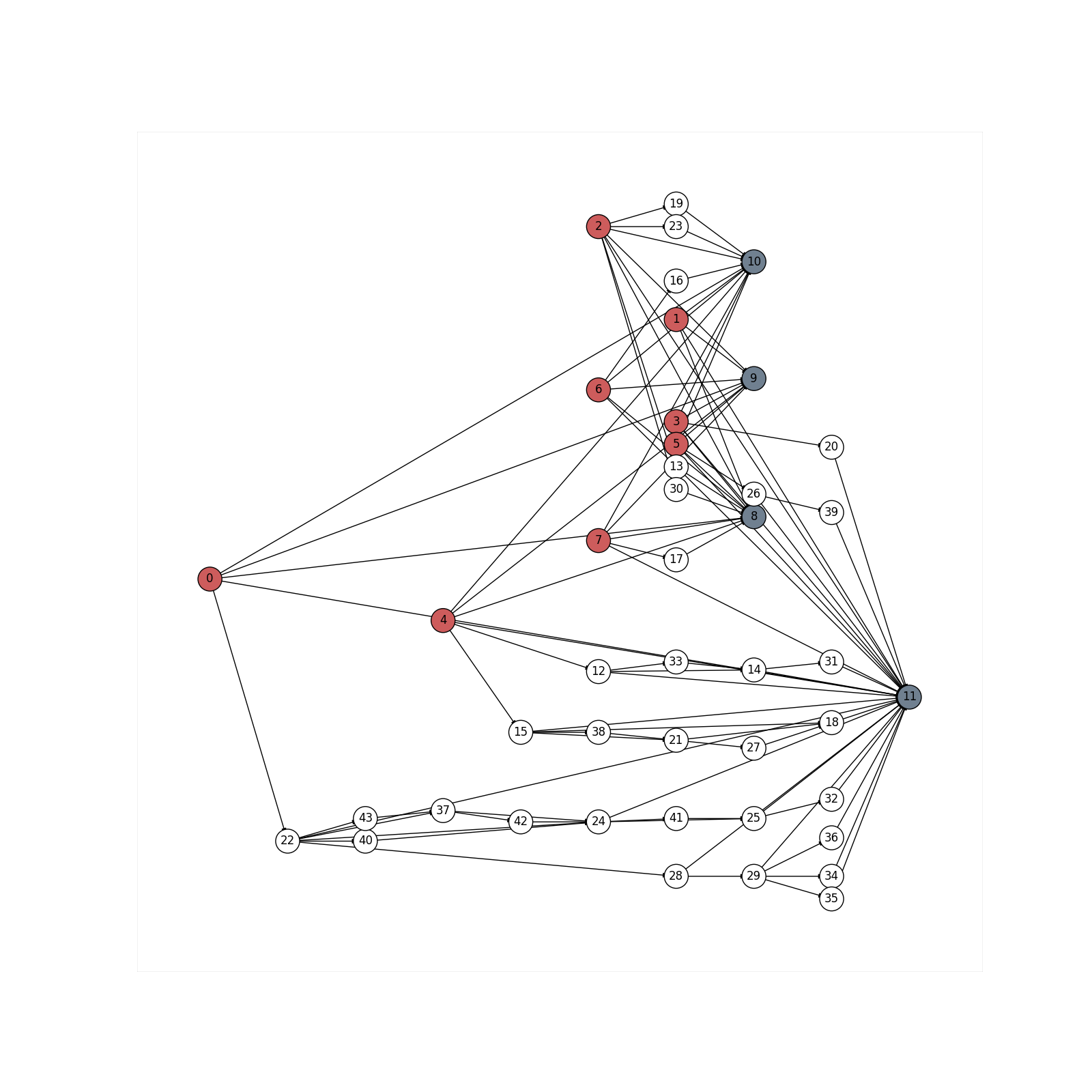}
         \caption*{64 growth steps. \\ 45 Nodes.}
         \label{fig:five over x}
     \end{subfigure}
    \caption{\textbf{Differentiable $\mathcal{NDP}$}: Lunar Lander Network policy growth over 64 steps. Red nodes are input nodes, blue nodes are output nodes, white nodes are hidden nodes.}
    \label{fig:lunar_example_diff_NDP}
\end{figure*}

\begin{figure*}[ht!]
     \centering
     \begin{subfigure}[b]{0.19\textwidth}
         \centering
          \captionsetup{justification=centering}
         \includegraphics[width=\textwidth]{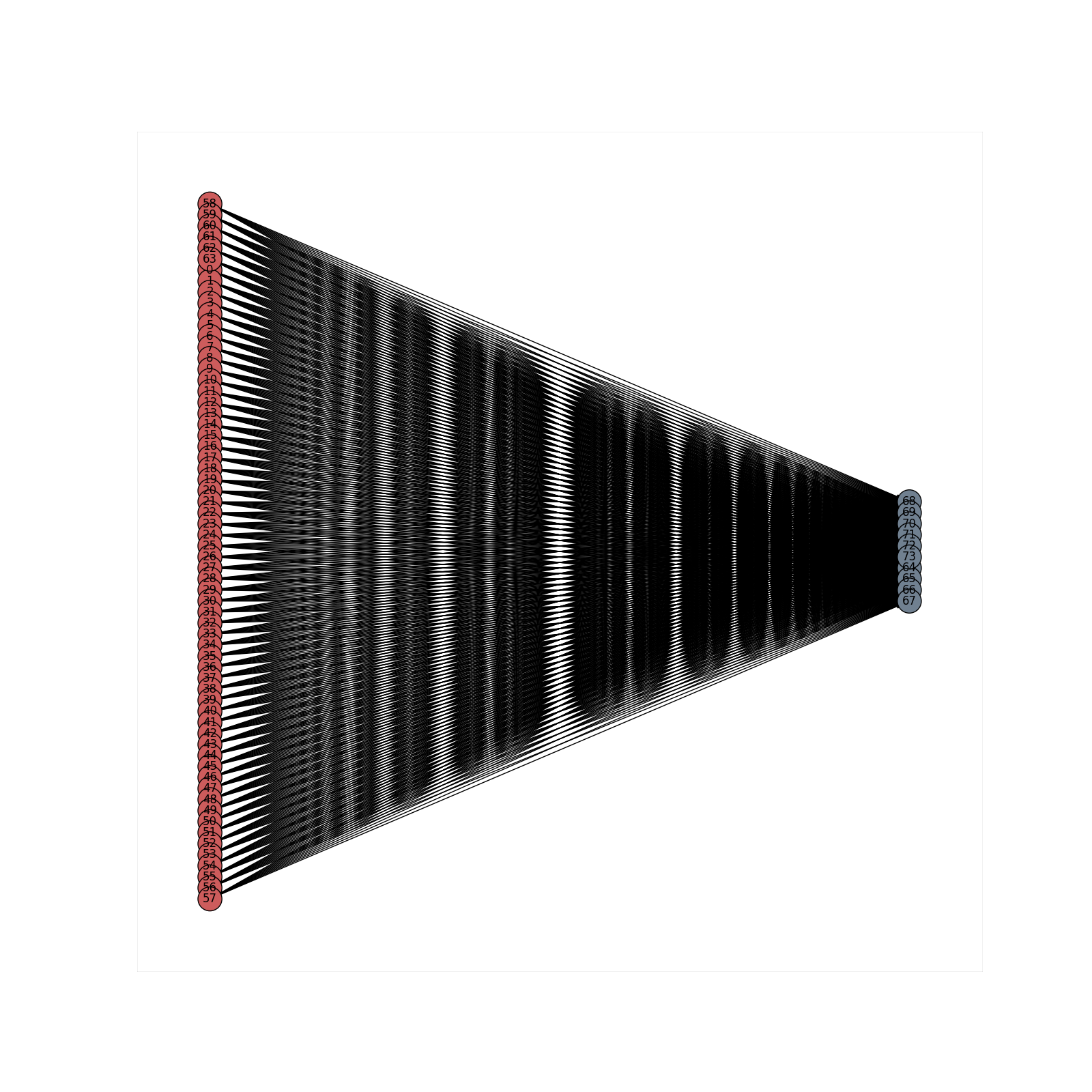}
         \caption*{Initial seeding graph. \\74 Nodes.}
         \label{fig:y equals x}
     \end{subfigure}
     \hfill
     \begin{subfigure}[b]{0.19\textwidth}
         \centering
         \captionsetup{justification=centering}
         \includegraphics[width=\textwidth]{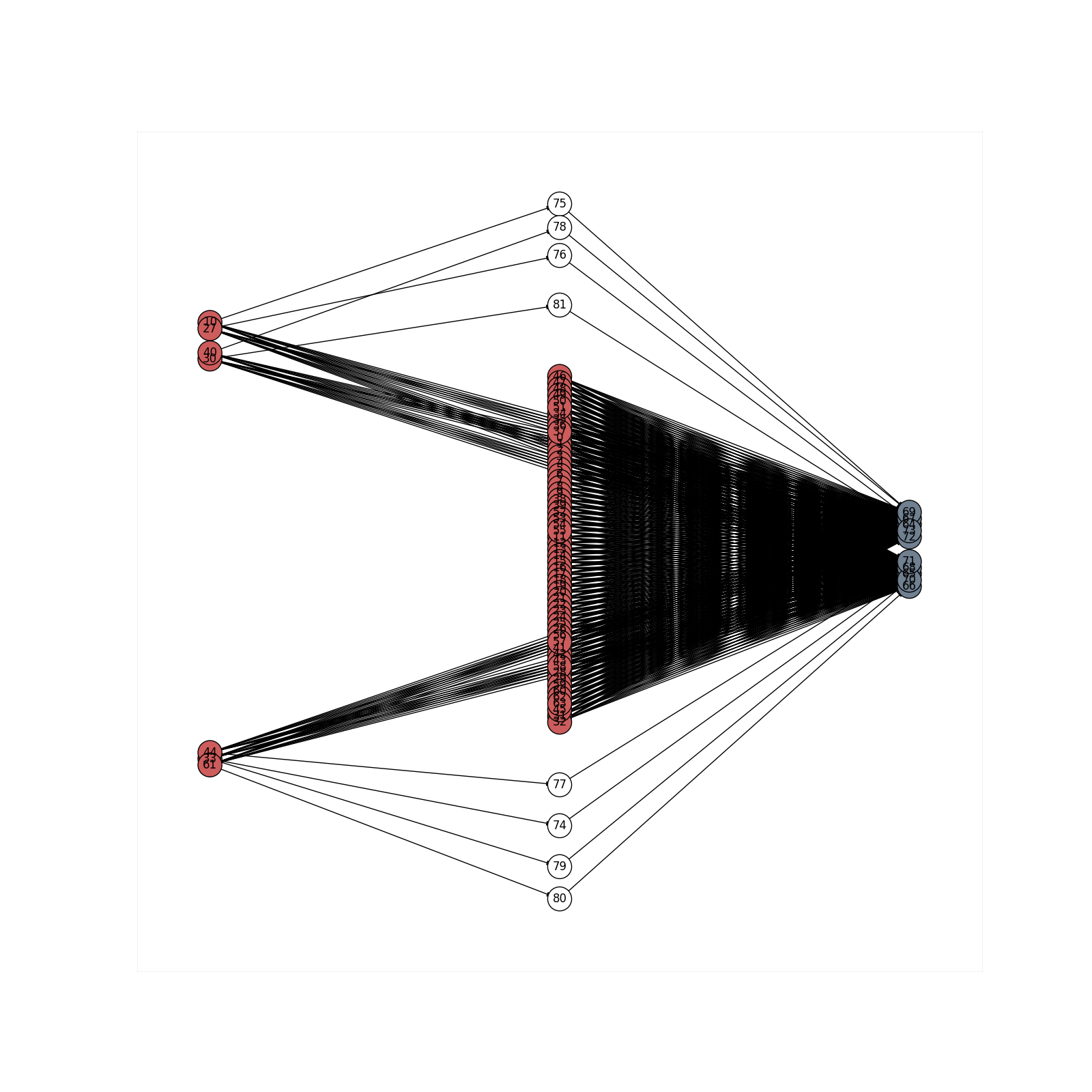}
         \caption*{16 growth steps. \\82 Nodes.}
         \label{fig:three sin x}
     \end{subfigure}
     \hfill
     \begin{subfigure}[b]{0.19\textwidth}
         \centering
         \captionsetup{justification=centering}
         \includegraphics[width=\textwidth]{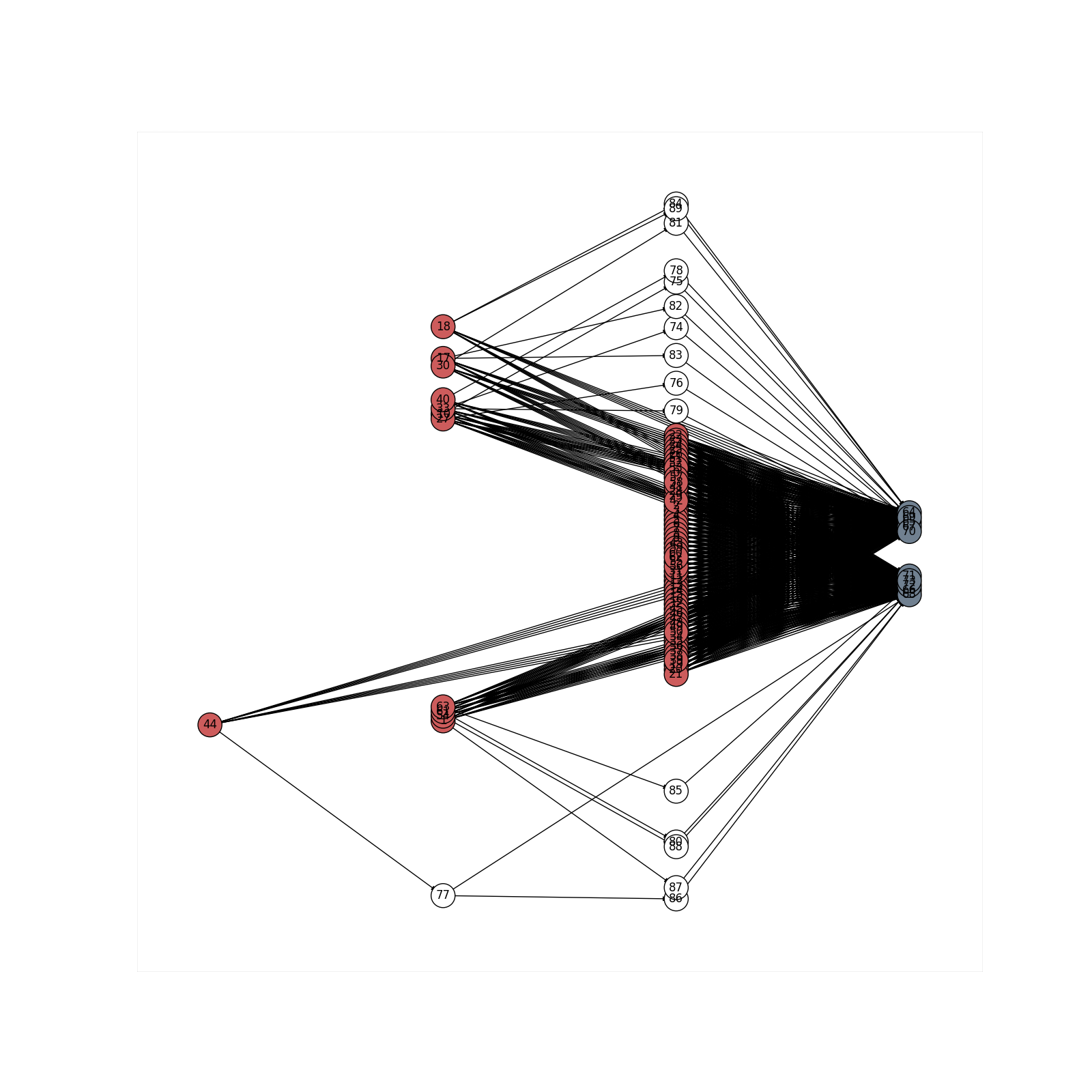}
         \caption*{32 growth steps.\\ 90 Nodes.}
         \label{fig:five over x}
     \end{subfigure}
     \hfill
     \begin{subfigure}[b]{0.19\textwidth}
         \centering
         \captionsetup{justification=centering}
         \includegraphics[width=\textwidth]{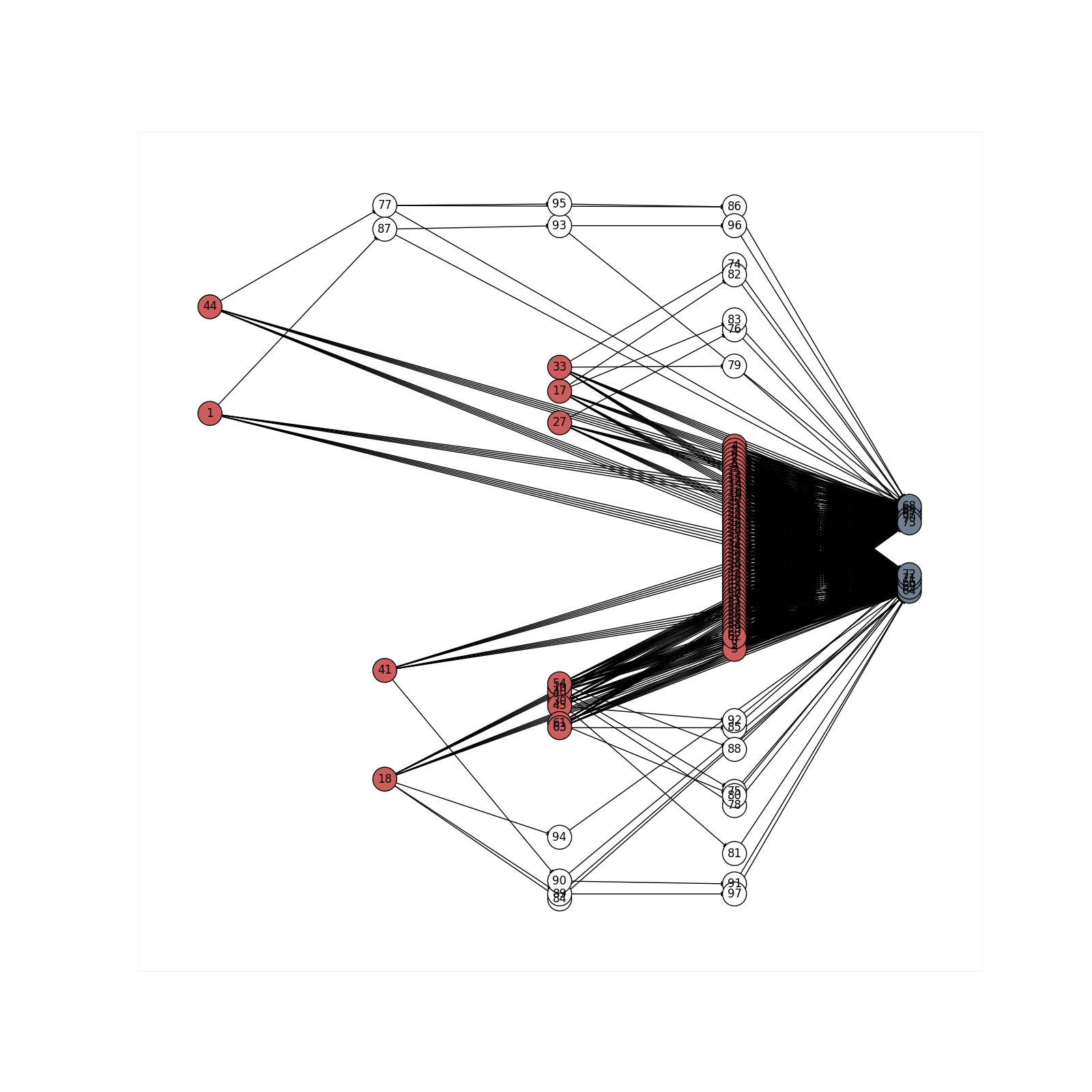}
         \caption*{48 growth steps. \\98 Nodes.}
         \label{fig:five over x}
     \end{subfigure} 
     \hfill
     \begin{subfigure}[b]{0.19\textwidth}
         \centering
         \captionsetup{justification=centering}
         \includegraphics[width=\textwidth]{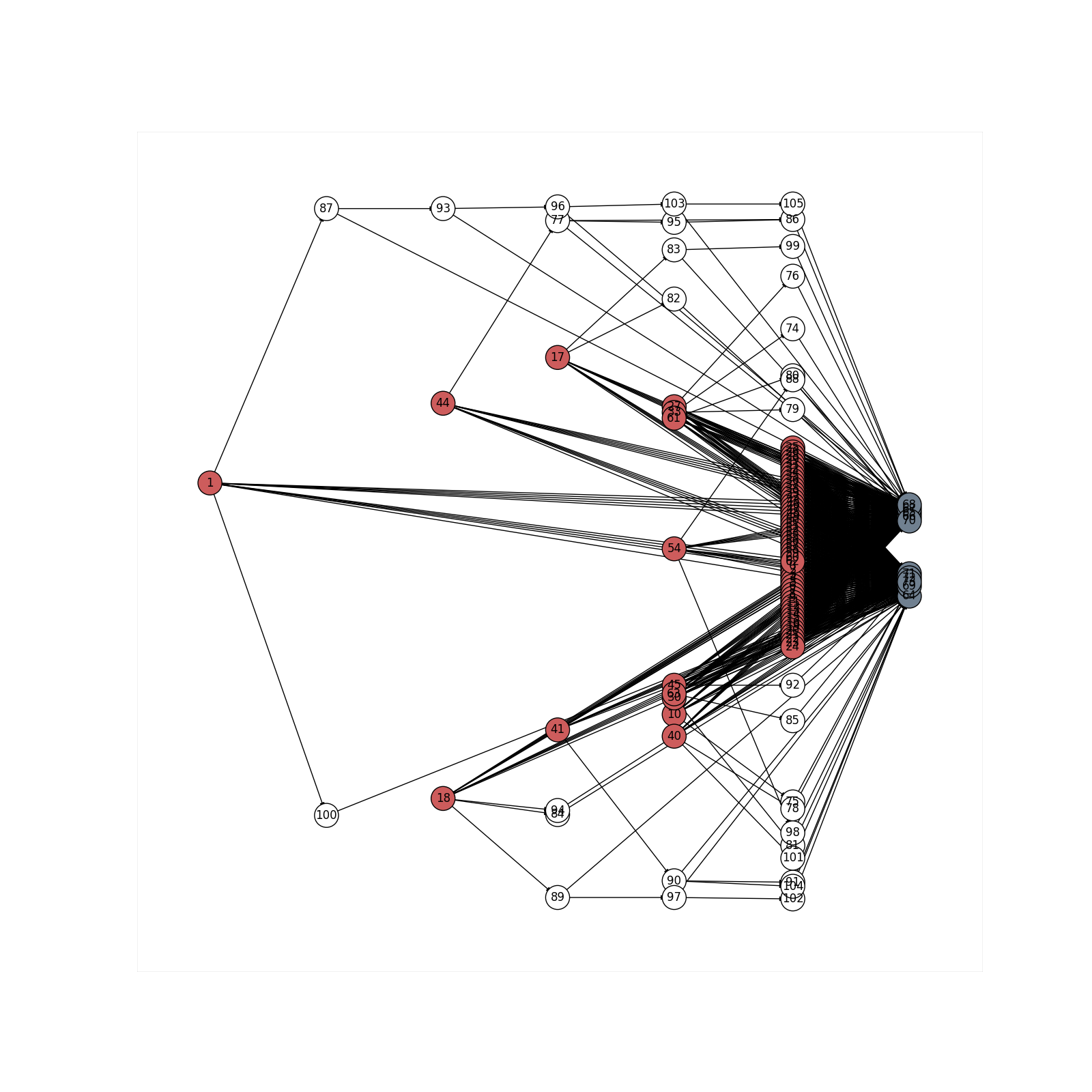}
         \caption*{64 growth steps.\\ 106 Nodes.}
         \label{fig:five over x}
     \end{subfigure}
    \caption{\textbf{Differentiable $\mathcal{NDP}$}: MNIST network growth over 64 steps.}
    \label{fig:mnist_growth}
\end{figure*}

We evaluate the differentiable $\mathcal{NDP}$ by comparing models that are trained and tested on different numbers of growth steps ("Growth Steps" column in Table~\ref{tab:onlineRL}). It seems that for most tasks, after a certain number of growth steps, the grown network's performance can deteriorate, as policies do not really benefit from more complexity. This could also be due to the simplicity constraint of grown architectures, making it unable to take advantage of new nodes as the networks get larger. Automatically learning when to stop growing will be an important addition to the $\mathcal{NDP}$ approach.

The differentiable $\mathcal{NDP}$ reaches comparable performance to the evolutionary-based version on CartPole (Table~\ref{tab:cartpole_grad}) and LunarLander (Table~\ref{tab:lunar_grad}). An example of the growth steps for the LunarLander tasks is shown in Figure~\ref{fig:lunar_example_diff_NDP}. In the offline RL setting, the NDP is able to get a mean reward of 29 on the HalfCheetah task (Table~\ref{tab:haflcheetah_grad}), which is satisfactory but lower compared to the performance of 43.1 achieved by behavioral cloning  in \citep{chen2021decision}. 


We are also able to scale up to larger networks for tasks like MNIST, which uses an 8$\times$8 image and reaches a test accuracy of 91\% (Table~\ref{tab:mnist_grad}).  The sequence of an MNIST network growing is shown in Figure~\ref{fig:mnist_growth}. 



\section{Discussion and Future Work}
We introduced the idea of a neural developmental program and two instantiations of it, one evolutionary-based and one gradient descent. We showed the feasibility of the approach in continuous control problems and in growing topologies with particular properties such as small-worldness. 
While the approach is able to solve these simple domains, many  future work directions remain to be explored. 

For example, the current $\mathcal{NDP}$ version does not include any activity-dependent growth, i.e.\ it will grow the same network independently of the incoming activations that the agent receives during its lifetime. However, biological nervous systems often rely on both activity and activity-independent growth; activity-dependent neural development enables biological systems to shape their nervous system depending on the environment. Similar  mechanisms also form the basis for the formation of new memories and learning. In the future, we will extend the $\mathcal{NDP}$ with the ability to also incorporate activity-dependent and reward-modulated growth and adaptation. 

While a developmental encoding offers the possibility to encode large networks with a much smaller genotype, the $\mathcal{NDP}$ in this paper are in fact often larger than the resulting policy networks. However, by running the developmental process longer, it is certainly possible to ultimately grow  policy networks with a larger number of parameters than the underlying $\mathcal{NDP}$. However, as the results in this paper suggest, growing larger policy networks than necessary for the tasks can have detrimental effects (Table~\ref{tab:lunar_grad}), so it will be important to also learn when to stop growing. The exact interplay between genome size, developmental steps, and task performance constitutes important future work. 

We will additionally extend the approach to more complex domains and study in more detail the effects of growth and self-organization on the type of neural networks that evolution is able to discover. $\mathcal{NDP}s$ offer a unifying principle that has the potential to capture many of the properties that are important for biological intelligence to strive \citep{versace2018priors, Kudithipudi2022nature}. 
While innate structures in biological nervous systems have greatly inspired AI approaches (e.g.\ convolutional architectures being the most prominent), how evolution discovered such wiring diagrams and how they are grown through a genomic bottleneck are questions rarely addressed.  In the future, $\mathcal{NDP}s$ could consolidate a different pathway for training neural networks and lead to new methodologies for developing AI systems, beyond training and fine-tuning.




\section*{Acknowledgments}
This project was supported by a GoodAI Research Award and a  European Research Council (ERC) grant (GA no. 101045094, project "GROW-AI")

\footnotesize
\bibliographystyle{apalike}
\bibliography{references}

\end{document}